%% file: main.tex
\documentclass[preprint,12pt]{elsarticle}

\usepackage[acronym]{glossaries}
\newacronym{genai}{GenAI}{Generative Artificial Intelligence}
\newacronym[plural=LLMs]{llm}{LLM}{Large Language Model}
\newacronym{nlp}{NLP}{Natural Language Processing}
\newacronym{cot}{CoT}{Chain-of-Thought}

\usepackage{amssymb}
\usepackage{amsmath}

\usepackage{xcolor}

\usepackage{hyperref}

\usepackage{tablefootnote}

\usepackage{subcaption}

\journal{Int. Journal of AI in Education}

\begin{document}

\begin{frontmatter}



\title{On the effectiveness of LLMs for automatic grading of open-ended questions in Spanish}



\author[label1,label2]{Germ\'an Capdehourat}
\author[label1,label2]{Isabel Amigo} 
\author[label1,label3]{Brian Lorenzo}
\author[label1]{Joaqu\'in Trigo}

\affiliation[label1]{organization={Ceibal},
            addressline={Av. Italia 6201, Edificio Los Ceibos}, 
            city={Montevideo},
            postcode={11600}, 
            country={Uruguay}}

\affiliation[label2]{organization={Eng. School, UdelaR},
    addressline={Av. Julio Herrera y Reissig 565}, 
    city={Montevideo},
    postcode={11300}, 
    country={Uruguay}}

\affiliation[label3]{organization={Eng. School, UCUDAL},addressline={Av. 8 de Octubre 2738}, city={Montevideo},postcode={11600}, country={Uruguay}}

\begin{abstract}
Grading is a time-consuming and laborious task that educators must face. It is an important task since it provides feedback signals to learners, and it has been demonstrated that timely feedback improves the learning process. In recent years, the irruption of \glspl{llm} has shed light on the effectiveness of automatic grading. In this paper, we explore the performance of different \glspl{llm} and prompting techniques in automatically grading short-text answers to open-ended questions. Unlike most of the literature, our study focuses on a use case where the questions, answers, and prompts are all in Spanish. Experimental results comparing automatic scores to those of human-expert evaluators show good outcomes in terms of accuracy, precision and consistency for advanced \glspl{llm}, both open and proprietary. Results are notably sensitive to prompt styles, suggesting biases toward certain words or content in the prompt. However, the best combinations of models and prompt strategies, consistently surpasses an accuracy of 95\% in a three-level grading task, which even rises up to more than 98\% when the it is simplified to a binary right or wrong rating problem, which demonstrates the potential that \glspl{llm} have to implement this type of automation in education applications.
\end{abstract}




\end{frontmatter}



\input{intro_and_soa}

\input{dataset_desc}

\input{experimental_setup}

\input{results_and_findings}

\input{ethics_conc_fw}



\bibliographystyle{elsarticle-num} 
\bibliography{llms_auto_grading}










\appendix
\include{prompts}

\end{document}

%% file: intro_and_soa.tex
\section{Introduction}\label{sec:intro}

The irruption of \glspl{llm} in recent years is redefining the way we interact with computational systems, as well as challenging many aspects of our daily lives, in several fields. Education is without a doubt one domain that is highly impacted by the capacities of \glspl{llm}, and moreover, by their widely accessible characteristic. In particular \glspl{llm} and, more generally \gls{genai}, are believed to have the potential to help solve many of the educational challenges of the 21st century. The possibility of implementing solutions on a large scale and at reduced costs \cite{USEd}, enables to develop novel tools and pedagogical innovation which paves the way towards the achievement of Sustainable Development Goals (SDG) number 4 \cite{Miao2021}. In particular, some examples are the possibility to adapt learning resources to the needs and strengths of each student, a higher customization of educational resources in order to be adapted to the geographical context, and the support to teachers and to the educational system in terms of automated tools and recommendation tools, among others \cite{USEd}. 

In this context, several studies have addressed the potential and challenges of \glspl{llm} in education, see, for instance, \cite{Wang2024}. From impact assessment to practical implementations, the literature in the field is vast. Recently in \cite{Deng2024}, authors have gathered insights from several studies concluding positive short-term results by using ChatGPT for learning in higher education. In another vein, in \cite{Zhang2024}, \glspl{llm} are used to simulate classroom environments, which in turn can be a powerful tool to promote educational research, help design solutions, and assess their impact. 

Another broad category is constituted by practical implementations or tools for educational stakeholders, namely for students, for teachers, or for the educational system, such as schools or decision-makers. In this regard, use cases can range from pedagogical to managerial. See for instance, \cite{Yan2024}, where authors thoroughly review the use of \glspl{llm} to  automate and support educational tasks.

 Several of those use cases fall on the category of automation that save time for teachers, including the time-consuming task of assessing or grading assignments. In addition, automating grading can also lead to providing more meaningful and timely feedback, and allow to scale-up different assessment methods, which altogether may lead to more effective learning experiences \cite{Gaytan2007,Poulos2008}. 

In particular, the use of \glspl{llm} for automatic scoring or grading has been put forward in \cite{Jeon2023}. Moreover, in recent years, experimental approaches have shown the utility of \glspl{llm} in providing support for automating grading, as well as pointed out remaining challenges, as we show in Section \ref{sec:soa}. 

Automated assessment is not novel as a research field; see, for instance, \cite{Valenti2003} for a review of \gls{nlp}-based algorithms for text auto-scoring and commercial products prior to the irruption of \glspl{llm}. However, \glspl{llm} shed new light on auto-scoring assignments of different natures, such as math, code, images, text, and language, as they are widely available, pre-trained systems that excel in their capability to provide convincing feedback. Nevertheless, their effectiveness remains a hot research field.

In this context, most of the literature concentrates on applications for the English language, and there are few works in languages other than English.
The present research work is focused on the performance analysis using \glspl{llm} for automatic grading of short-textual open-ended questions in Spanish. 
It also stands out by considering the prompts in the same language (i.e. Spanish), which we believe is a key aspect for this type of application, since it is natural to be the teachers who should define the prompts, and do them in the same language as the task to be graded. It is important to clarify that we are not analyzing an scenario where Spanish is taught as a second language, but rather it is the native language in the context under study.

In the next section we present a brief summary of related works in the area of \gls{genai} and \glspl{llm} education applications such as automated grading. By means of a handcrafted dataset described in Section~\ref{sec:dataset}, which includes  human-expert ratings as ground truth, several combinations of state-of-the-art \glspl{llm} and prompting strategies were evaluated. The corresponding experimental setup is depicted in Section~\ref{sec:experimental_setup}. In addition to analyzing the comparative accuracy of the different options, experiments were also carried out to study the grading consistency and the sensitivity to changes in the prompts. The main findings from the experiments, detailed in Section~\ref{sec:results_and_findings}, indicate that, while a significant impact is observed from the specific words used in the prompts, the high capacity \glspl{llm} combined with few-shot based prompting (which turned out to be the best option) are able to consistently achieve a very high accuracy for the automated grading task analyzed. Finally, conlusions and future work are discussed in Section~\ref{sec:concl_and_fw}, while Section~\ref{sec:ethics} focuses on ethics aspects which are beyond this study. We also include~\ref{sec:app_prompts} detailing all the prompts used in the experiments.

\section{Related Work} \label{sec:soa}
The literature on \glspl{llm} and, more broadly, \gls{genai} for automatic grading is both recent and extensive. In this section, we focus on studies that demonstrate different approaches to achieving effective results in auto-grading assignments of various types.


In \cite{Mohamed2025} authors show the effectiveness of \glspl{llm} for evaluating code assignments. In particular, they carry out experiments comparing their performance with human evaluators across various programming assignments, and different \glspl{llm}, concluding that advance \glspl{llm} highly correlate with human evaluators.

In \cite{Latif2024_distilation}, a specific method called knowledge distilling is proposed to further improve the performance of a fine-tuned \gls{llm} in science assessments automatic grading, showing promising results on using GenAI for such purposes.

Several studies focus on the evaluation of language assignments through short textual answers to open-ended questions or through essays, mostly in English.  That is the case in \cite{Chu2024} where a multi-agent system based on \glspl{llm} is proposed, or in \cite{Xie2024} where a complete assessment system is proposed, including elaboration of rubrics, prompt creation based on those rubrics, task assessment and output evaluation for system improvement. 

In a recent study \cite{Schneider2024}, authors focus on a setting for auto grading short textual answers. On the one hand, they conclude that if the LLM-based approach for such task is promising, full automation is not feasible nowadays. On the other hand, they highlight the impact of the language in use on results, showing different effectiveness in answers provided in English than in German. 

Also related to English automated assessment, in this case of essays, \cite{Caines2023} concludes that while \glspl{llm} show potential for automated grading, they do not outperform traditional methods that use established linguistic features on their own. In \cite{Stahl2024} authors study several prompting strategies to provide jointly automate scoring and feedback to essay writing in English. Their results suggest that auto scoring can benefit feedback generation and vice versa, and that good results can be achieved by prompting. In particular, they compare different prompts responding each to a different profile of evaluator. In addition, different combinations of scoring and feedback are asked as tasks to the \gls{llm}. 

In terms of texts written in other languages, research on automatic scoring is clearly less abundant. In \cite{Chang2024}, authors evaluate the use of ChatGPT in Finnish short answers, obtaining good results with advanced \glspl{llm} versions. In \cite{Mardini2024}, authors study the use of BERT for automating short answer grading in Spanish. More recently, authors in \cite{Capdehourat2024} focus on the effectiveness of \glspl{llm} for the assessment of answers of open-ended questions in Spanish showing promising results. Our work extends those results by thoroughly investigating, through an experimental approach, additional language models and prompting strategies. In addition, we focus on relevant aspects such as the \glspl{llm} consistency and performance robustness against subtle changes in the prompts.

Our literature review has shown that the performance of \glspl{llm} for automatic grading varies according to several parameters, including the model itself, the task it is asked to assist with, the prompt, the language being used, and the techniques employed to refine the model, such as fine-tuning, few-shot learning, chain of thought, mixture of experts, or multi-agent-based solutions. Additionally, to the best of our knowledge, little literature exists on the specific use case where the language in use is Spanish, whether for prompting, for the text being graded, or for providing feedback.

 In the remainder of this paper, we focus on the automatic grading of short answers to open-ended questions in Spanish through an empirical approach. We shall recall that the use case under analysis corresponds to a context where Spanish is the mother tongue. It is essential that the prompts used are in the same language, as we believe teachers should take the lead in defining these prompts when automated task grading solutions are adopted on a large scale.

%% file: dataset_desc.tex
\section{Distilled Dataset for the Open-ended Questions Use Case}
\label{sec:dataset}

The proposed use case targets students in the middle stage of primary education, that is to say, students ranging from 8 to 10 years old. This context emphasizes fostering critical thinking and expressive communication skills through interactive question-answering tasks. To cover a diverse array of topics and ensure comprehensive engagement, we build a dataset including several categories such as personal background and family composition, likes and dislikes, preferences and favorites, academic interests, daily routines, and recreational activities. A collection of 51 meticulously crafted questions serves as the foundation for this dataset, each designed to elicit nuanced responses. 

For every question, three tiers of sample answers were created: \emph{excellent}, reflecting a complete and insightful response; \emph{acceptable}, representing a satisfactory but less elaborate reply; and \emph{incorrect}, capturing common errors or misconceptions. The following subsections elucidate the systematic approach undertaken to curate this dataset, ensuring its relevance and applicability for educational research and development.

\subsection{Dataset Creation Procedure}

The initial step involved selecting a set of 51 human-written questions in Spanish, evenly distributed across the predefined topic categories. These questions served as the foundation for generating a diverse range of answers using a state-of-the-art language model. Specifically, we employed GPT-4\footnote{The GPT-4 version used was \texttt{gpt-4-turbo-2024-04-09}.} from OpenAI with default parameters (i.e., temperature = 1 and maximum output tokens = 256). After iterative testing with different prompt structures, a final version was crafted to systematically generate responses for the three quality tiers:
\begin{itemize}
    \item \textbf{Excellent}: Comprehensive and well-justified answers enriched with relevant details.
    \item \textbf{Acceptable}: Correct responses that align with the question but provide minimal elaboration.
    \item \textbf{Incorrect}: Answers that deviate from the instructions and are deemed inaccurate or irrelevant.
\end{itemize}

To ensure reliable outcomes with GPT-4, we found it necessary to limit the number of responses requested in a single run. For each question, 4 responses of each type were generated per run, and this process was repeated twice. Consequently, a total of 24 responses were generated for each question, comprising 8 excellent, 8 acceptable, and 8 incorrect answers. Both the prompts and the generated responses were consistently in Spanish, maintaining linguistic and contextual integrity throughout the dataset creation process.

Finally, a thorough human-based quality control process was conducted to refine the GPT-4-generated responses. While the overall quality was high, manual corrections were necessary for certain entries. Of the 1,224 automatically generated answers, 227 (18.5\%) required adjustments. Most corrections addressed repeated or overly generic incorrect answers, while others involved rectifying typographical or grammatical errors. Additionally, a few cases presented inconsistencies, such as responses labeled as \emph{incorrect} that could be considered \emph{acceptable} upon closer inspection.

The following illustrates one of the dataset questions along with representative examples of responses across the three quality tiers:

\begin{center}
\noindent
\fcolorbox{black}{lightgray}{\parbox{0.95\linewidth}{{\bf Question:} ?`Cu\'al es tu mascota favorita?\\

{\bf Example Answers:}
\begin{itemize}
    \item ``Los h\'amsters son mis favoritos porque son peque\~nos, f\'aciles de cuidar y me divierte verlos correr en sus ruedas.'' - Excellent.
    \item ``Prefiero los conejos.'' - Acceptable.
    \item ``El mejor d\'ia de la semana es el viernes.'' - Incorrect.
\end{itemize}
}}

\end{center}

The complete dataset, along with all the prompts employed in this study, is openly available on GitHub: \textcolor{gray}{Link omitted for double-blind review}. 

%% file: experimental_setup.tex
\section{Experimental Setup}\label{sec:experimental_setup}

In this section, we present the experimental framework established to evaluate the performance of various language models. This framework comprises three core elements: the selection of language models (\glspl{llm}), the design of various prompting strategies, and the description of the software and hardware utilized to execute the experiments.

\subsection{Selected Models}
\label{sec:models}

A range of language models, encompassing both proprietary and open-weight solutions, was chosen for the experiments. For proprietary models, we utilized OpenAI's offerings, specifically GPT-4o and GPT-4o-mini. The selection of open models was guided by performance metrics from sources such as Open \gls{llm} Leaderboard~\cite{llm_leaderboard} and the LMSYS Chatbot Arena~\cite{Chiang2024}, as well as availability through platforms such as \textit{Ollama}\footnote{Ollama: \url{https://www.ollama.com/}}.
The open models selected in this case were Gemma2 by Google~\cite{gemma2}, Llama3 by Meta~\cite{llama3} and Qwen2.5 by Alibaba~\cite{qwen2.5}.

All the selected models and their configurations are detailed in Table~\ref{tab:models}, which provides a comprehensive overview of their parameters and capabilities. The number of parameters of proprietary models was based on the estimations included in this paper~\cite{Abacha2025}, as there is no public information from OpenAI.

 This diverse selection aims to encompass a wide spectrum of model sizes and capabilities, representative of the current state of the art in \glspl{llm}. While many other alternatives could have been considered, both proprietary and open, we believe this curated set offers a balanced and comprehensive basis for evaluating performance in the specific context of the automated grading task under study.

\begin{table}[]
\begin{center}
\renewcommand{\arraystretch}{1.1} 
\begin{tabular}{lllll}
\hline
\multicolumn{1}{c}{Model} & \multicolumn{1}{c}{Provider} & \multicolumn{1}{c}{\begin{tabular}[c]{@{}c@{}}Number of\\ parameters\end{tabular}} & \begin{tabular}[c]{@{}l@{}}Context\\ length\end{tabular} & \multicolumn{1}{c}{License} \\ \hline
Gemma2 - 2B & Google & 2B      & 8k  & Open         \\
Llama 3.2 - 3B & Meta          & 3B  & 128k   & Open         \\
Qwen2.5 - 7B & Alibaba       & 7B & 128k & Open         \\ 
GPT-4o-mini   & OpenAI        & $\thicksim$8B & 128k & Proprietary  \\
Gemma2 - 9B & Google & 9B      & 8k  & Open         \\
Llama 3.3 - 70B & Meta          & 70B  & 128k   & Open         \\
Qwen2.5 - 72B & Alibaba       & 72B  & 128k & Open         \\ 
GPT-4o  & OpenAI        & $\thicksim$200B     & 128k & Proprietary  \\
\hline
\end{tabular}
{Estimations for OpenAI models were taken from~\cite{Abacha2025}.}
\caption{List of selected models for the experiments.}
\label{tab:models}
\end{center}
\end{table}

\subsection{Prompting Techniques}
\label{sec:models}

In addition to testing with different LLMs, we also analyzed several prompting strategies to assess their impact on automated grading performance. Various prompts were designed following several general principles from the recent literature in the area (e.g., Section 4.2 of~\cite{Schulhoff2024} includes further details on prompting options for evaluation purposes). All of them indicate the \gls{llm} to use a three-level Likert scale for grading (i.e. excellent, acceptable, and incorrect). Additionally, it is worth to recall that all the prompts tested were in Spanish.

To start with, two base prompts were considered: one short concise version and a longer more detailed one. The first one simply outlined the three-level grading scale and asks the model to qualify the response. On the other hand, the more detailed prompt elaborates on the characteristics of each level (in a similar way as they were presented in section~\ref{sec:dataset}, but in Spanish). 
Thus, the extended prompt is intended to function effectively as a grading rubric.

Building upon the previously defined base prompts, we generated two new versions based on the \gls{cot} prompting strategy. The main idea behind \gls{cot} is to ask for the \Glspl{llm} to \emph{think step by step}, based on the empirical results that show use cases where including that in the prompt lead to better performance~\cite{Wei2024}. In our case, we included in the prompt the instruction to the \gls{llm} to provide feedback on the response to evaluate, before indicating the corresponding rating. Thus, the resulting prompts resemble a \gls{cot}-based approach, in terms of asking the models to thoroughly \emph{think} on the responses before grading them.

Finally, the last strategy considered was few-shots based prompting~\cite{Brown2020}. In this case we just used the short base prompt, but integrating also an open-ended question example that resembles the ones in the dataset, and several responses with their corresponding gradings. In this way, it is expected that \glspl{llm} by analogy will appropriately grade the new responses accordingly to the defined ratings scale.

To summarize, five different prompts were designed based on the strategies detailed above:
\begin{itemize}
    \item \textbf{Short Concise Prompt:} It includes the question and the answer to evaluate, and instructs the \gls{llm} to generate the grading according to the three-level scale. \texttt{Number of words:} \textbf{52}.
    \item \textbf{Long Detailed Prompt:} It adds to the short version the assignment of the role as a teacher, the context in primary education and a description of the grading scale. \texttt{Number of words:} \textbf{139}.
    \item \textbf{\gls{cot}-based Short Prompt:} It extends the short concise prompt by asking the LLM to provide feedback about the answer before assigning a grade. \texttt{Number of words:} \textbf{67}.
    \item \textbf{\gls{cot}-based Long Prompt:} It integrates the same extension as the previous strategy, but taking as base prompt the long detailed one. \texttt{Number of words:} \textbf{154}.
    \item \textbf{Few-Shot based Prompt:} It includes an example question and several responses with their corresponding gradings within the prompt. As in the previous short versions, it does not include a role assignment or a description of the context. \texttt{Number of words:} \textbf{202}.
\end{itemize}
\noindent It is worth clarifying that the number of words in each test execution will vary, since it depends on the specific question and answer to be evaluated. In this case, the \texttt{number of words} indicated for each prompt corresponds to the minimum each one will have, without counting what the variable parts add according to the specific input. As we can see, there are significant differences between prompts regarding the number of words, which is a good proxy to indicate the number of tokens for each case. This implies an important difference when comparing the resources needed to execute them, which we will discuss later in Section~\ref{sec:results_and_findings} when we discuss the results obtained.

Each of the \glspl{llm} was evaluated with all the defined prompts. 
These strategies were selected based on their potential to enhance the accuracy of \glspl{llm}, as suggested by recent advancements in prompt engineering research. The full text of the each prompt considered in the experiments is detailed in \ref{sec:app_prompts}.

\subsection{Software and Hardware for Running the Tests}

The testing framework leveraged \textbf{Ollama} alongside \textbf{Promptfoo}\footnote{Promptfoo: \url{https://github.com/promptfoo/promptfoo}}.
Ollama is a powerful open-source platform that enables running \glspl{llm} on local infrastructure. It supports most of the state of the art open models, which are quickly made available whenever a new one comes out. On the other hand, Promptfoo is a JavaScript-based tool designed for evaluating the quality of outputs generated by LLM-driven applications. Promptfoo provides a declarative testing environment through YAML configuration files, enabling the specification of various models, prompts, and assertion functions within a single test. These assertion functions compute performance metrics against the test dataset. Furthermore, Promptfoo seamlessly integrates with proprietary model providers such as OpenAI, requiring only a valid API key for execution.

All experiments were conducted on a server equipped with an Intel Xeon CPU E5-2680 v3 @ 2.50GHz, running Ubuntu 22.04.4 LTS. The server featured a 512GB SSD for storage and 96GB of RAM, along with an NVIDIA GeForce RTX 3090 Ti GPU with 24GB of memory. This hardware configuration ensured efficient processing of the diverse test scenarios. Executing all tests across the selected \glspl{llm} and prompting strategies required approximately 80 hours of computation time, demonstrating the robustness and scalability of the experimental setup.

%% file: results_and_findings.tex
\section{Analysis of Results and Main Findings
}\label{sec:results_and_findings}

In this section we present the different experiments carried out and discuss about the most important insights found.
For the automated grading task under study, the goal is that the model ratings match the ground truth human-validated labels of the dataset. Thus, for all tests, the metric we will be considering corresponds to accuracy, which is the percentage of correct ratings in each case.
In each subsection we will focus on different aspects, starting with a general comparison for all the \glspl{llm} and prompts considered, and then delving into those that gave better results in the first instance.

\subsection{Comparative Results for all LLMs and Prompts}

\begin{figure}
    \centering
    \includegraphics[width=\linewidth]{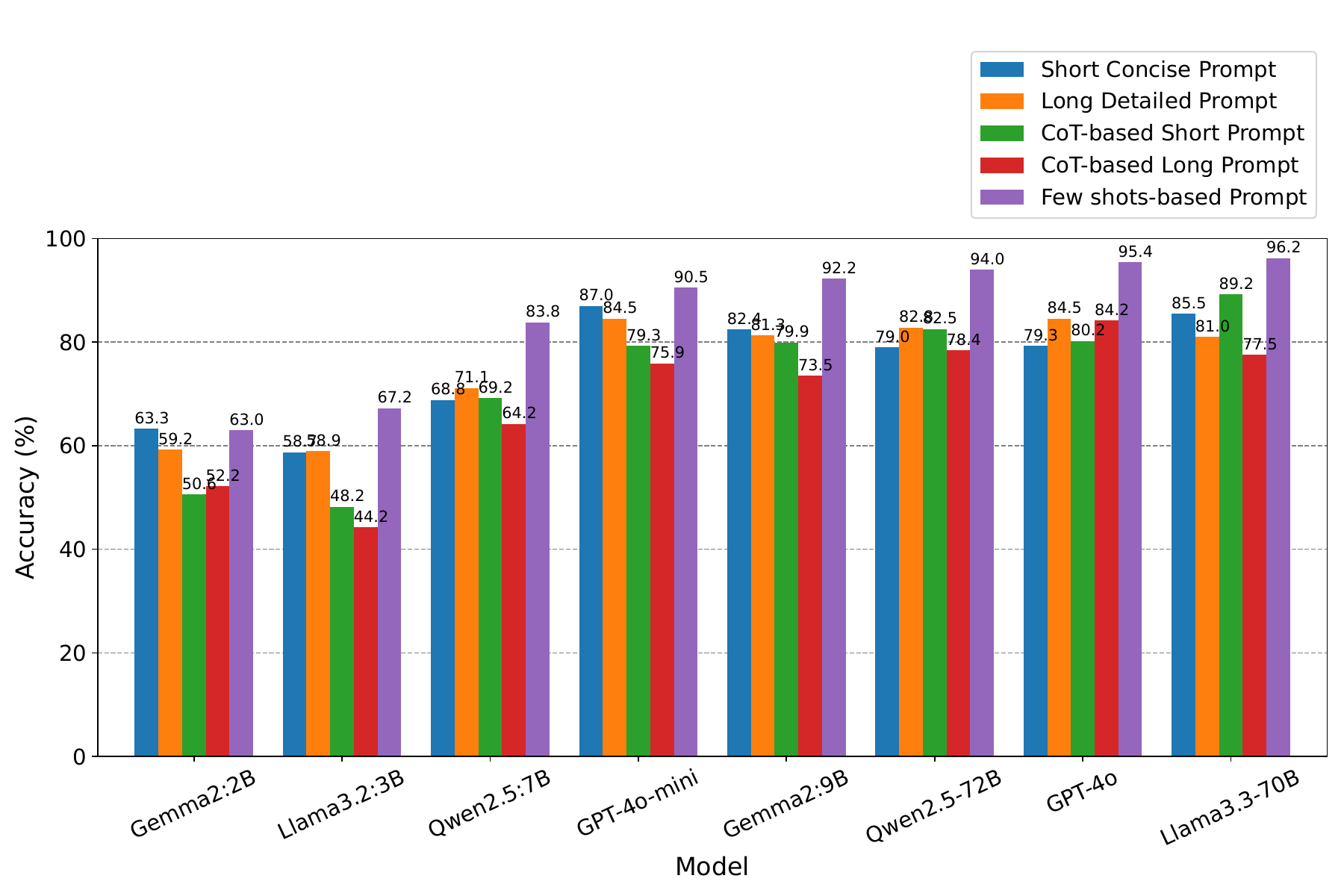}
    \caption{Performance results for the different models and prompts tested.}    
    \label{fig:accuracy_all}
\end{figure}

To begin with, we compare the performance of the different \glspl{llm} for all the prompt strategies presented previously. In Figure~\ref{fig:accuracy_all} all the results are shown. The first thing to notice is that perfomance increases as model capacities/sizes grow. It can be seen that the 80\% performance barrier is easily surpassed with some of the medium-sized models. Meanwhile, the best results are achieved with the large models, both open and proprietary, reaching an accuracy of over 95\% for the best cases.

Concerning the prompts, we have a clear winner which is the few-shot strategy. For all models it is the alternative with the best results, also showing a significant improvement in most cases compared to the other strategies. Among the latter, there is no clear efectiveness of any of them that leads to improvements in all cases. For example the more detailed prompts only show better results for GPT-4o, while the CoT-based approach only achieves some improvement for LLama3.3-70B.

To conclude this general comparison, we could summarize the results in three main takeaways. The first one is that the task of automatic grading of open-ended questions in Spanish seems feasible to be solved using state-of-the-art \glspl{llm}. It is important to remember that all the prompts used are also in Spanish.
Secondly, it should be used a medium to large model depending on the desired performance, requiring the latter if the goal is an accuracy greater than 95\%. Finally, with no doubt the best prompting strategy in terms of accuracy is few-shot, but we should not forget that this has the disadvantage of using a greater number of tokens. Approximately four times more tokens are needed than for the short concise prompt, thus increasing the resources/costs required for each automated grading request.

\subsection{Prompt Sensitivity Analysis}
\label{sec:prompt_sens}

Now that we have an overview of the results, the idea is to delve deeper into specific aspects. In this subsection we will focus on studying the sensitivity to changes in the prompt. The modification considered is something subtle but significant, which corresponds to replace the specific words that we will use for the three defined rating levels.

Since the prompts are in Spanish, so are the specific words used for the gradings. Initially, the ratings used were: \textsc{excelente}, \textsc{aceptable}, \textsc{incorrecta} (i.e. excellent, acceptable, incorrect). In this case, we repeated the same tests, but using the synonyms: \textsc{sobresaliente}, \textsc{bien}, \textsc{deficiente} (i.e. outstanding, good, poor). We shall call this two alternatives EAI ratings and SBD ratings, based on the initials of each respective case. 

\begin{figure}
    \centering
    \includegraphics[width=\linewidth]{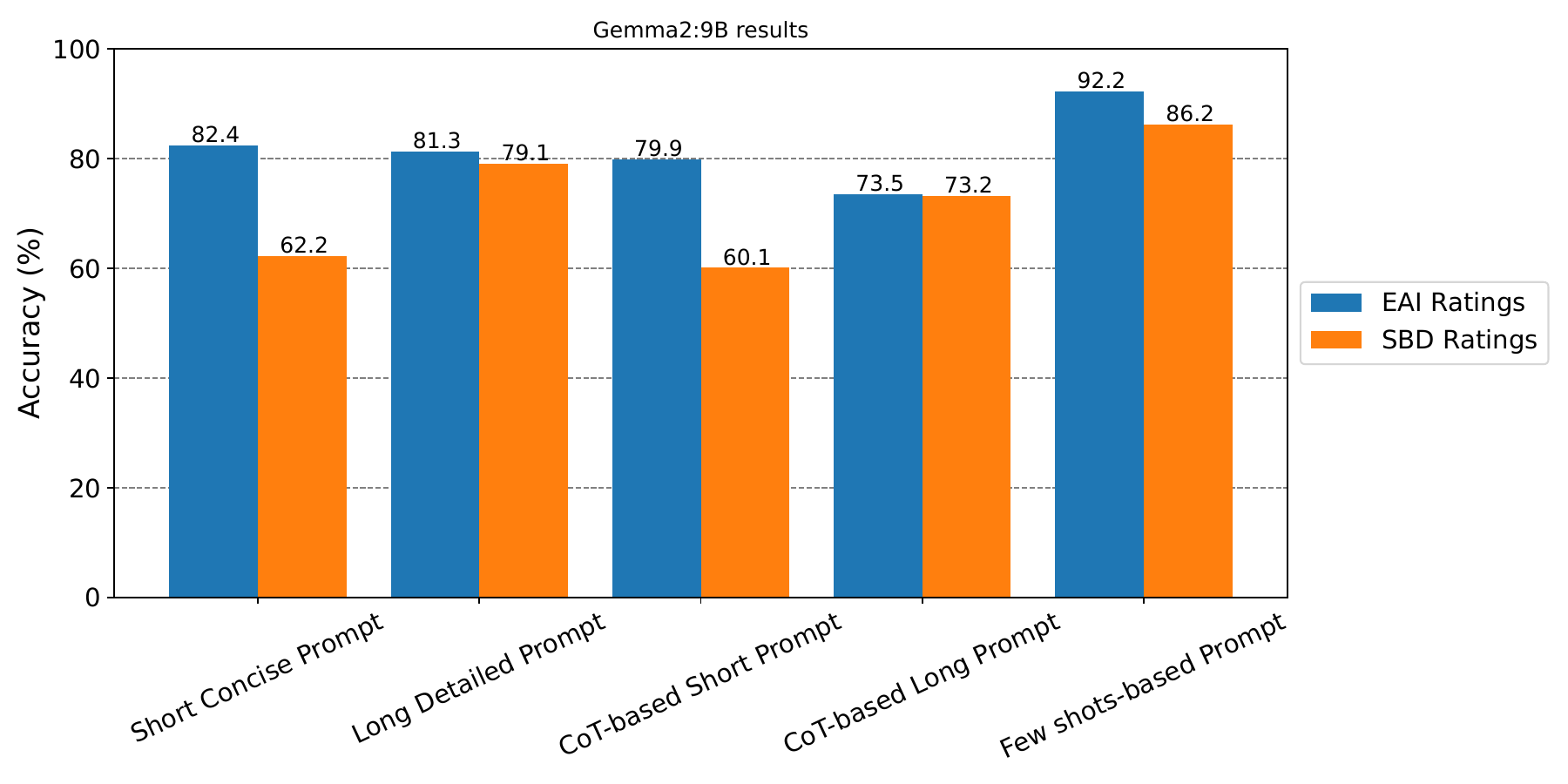}
    \caption{Prompt sensitivity analisys for the Gemma2-9B open model.}      
\label{fig:prompt_sens_gemma}
\end{figure}

The results show that, for all models and prompt strategies, this simple change in the words for the grading scale leads to worse results. Moreover, the performance drop is quite significative in most alternatives, reaching more than 10\% in some cases.
For example, in Figure~\ref{fig:prompt_sens_gemma} we observe the result of the open medium-sized model  Gemma2-9B, which presented pretty good accuracy in the first experiments. For the short prompts, where the ratings description is not included, the performance drop reaches 20\%. This implies that the model varies its behavior considerably depending on the specific words used as ratings. In the longer prompt versions, where the grading scale includes an associated rubric, it can be seen that the impact of changing the ratings words is much smaller.

\begin{figure}
    \centering
    \includegraphics[width=\linewidth]{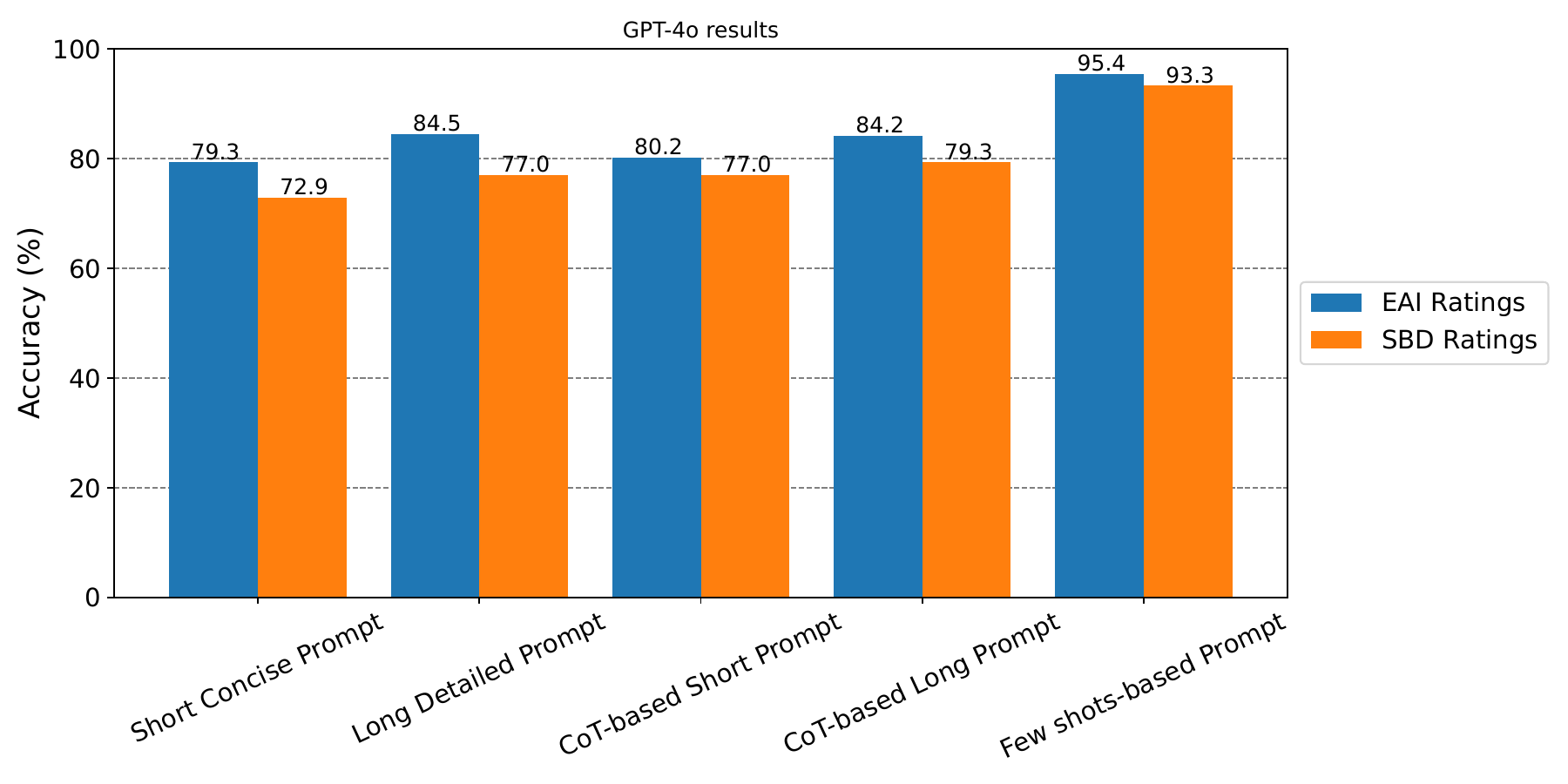}
    \caption{Prompt sensitivity analysis for the GPT-4o OpenAI model.}    
    \label{fig:prompt_sens_gpt4o}
\end{figure}

\begin{figure}
    \centering
    \includegraphics[width=\linewidth]{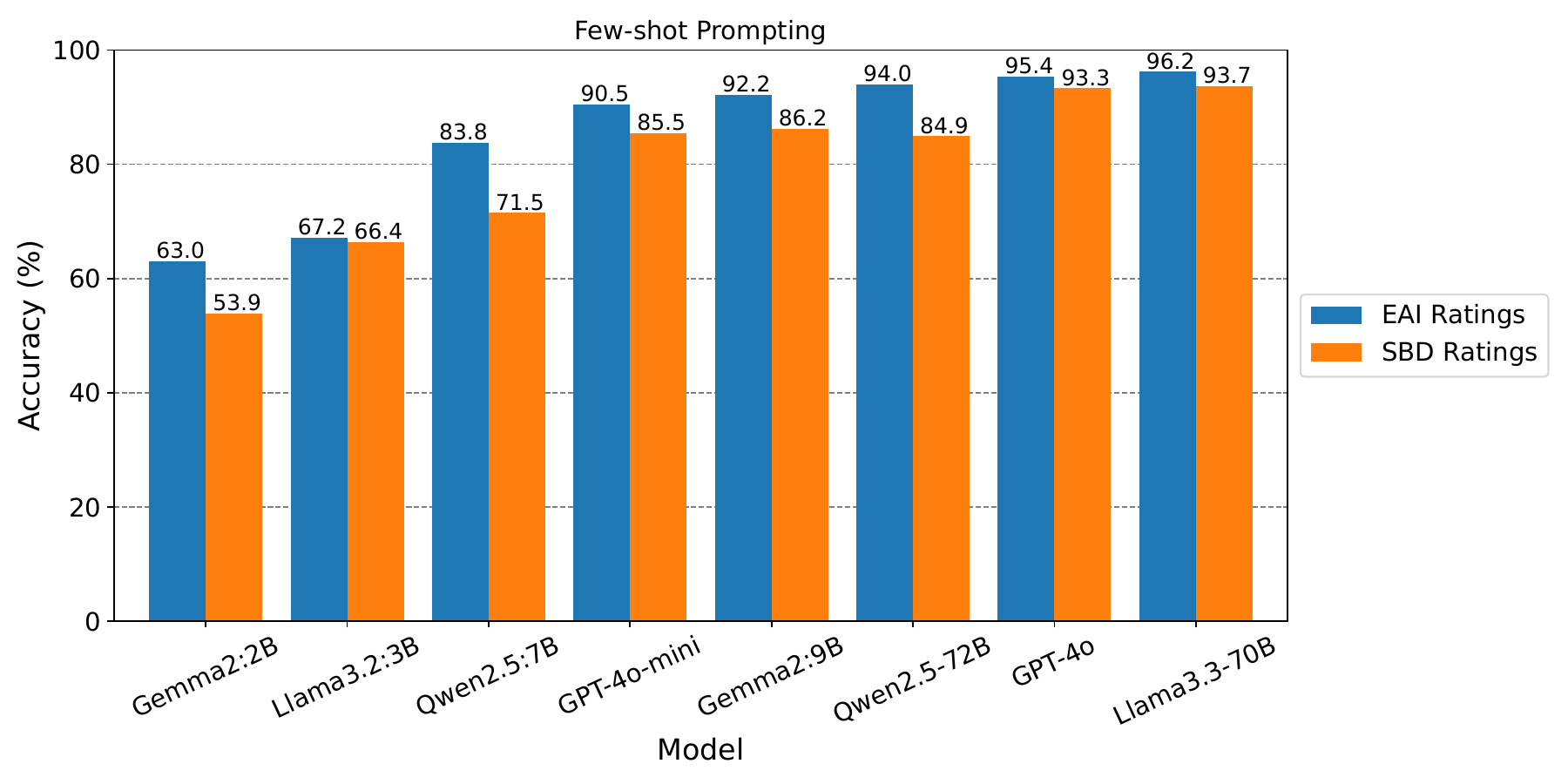}
    \caption{Prompt sensitivity analysis for all the models for the few-shot case.}    
    \label{fig:prompt_sens_all}
\end{figure}

On the other hand, in Figure~\ref{fig:prompt_sens_gpt4o} we observe the case of the GPT-4o model, where we can see a lower impact associated with this change, with few-shot being the most robust prompting technique. Comparing the results of all models using this prompting strategy (see Figure~\ref{fig:prompt_sens_all}) we can see that the impact is moderate except for a few exceptions. The performance drops range from 0,8\% to 12.3\%. In particular, for the best two models the impact is below 3\%, being 2.1\% for GPT-4o and 2.5\% for Llama3.3-70B.

The performance impact observed when a slight change in the particular rating words used in the prompt is a very important finding for the automatic grading task under study. 
This sensitivity of \glspl{llm} when dealing with subtle prompt changes could be a major drawback for this specific education application. We envision education tools based on \glspl{llm} where the prompts are written directly by teachers, detailing the particular grading rubric that should be used for a certain task. Thus, if LLMs are prone to performance drops related to the particular words used in the prompts, this could lead to major failures and grading biases generated by the specific writing styles of the different teachers.

\subsection{LLMs Consistency Analysis}

Another major issue when developing applications with \glspl{llm} is how to manage the randomness of their outputs. Although it is possible to work on this aspect, for example by regulating the input temperature parameter that with \glspl{llm} usually have, we decided to experiment a different approach. Our focus was on analyzing the variations that models have when using the default parameters, which is the case that most of the people may typically use.

In order to analyze the consistency of the results, we had to make several executions with the same question-answer input pairs from the dataset. Since doing this with the entire dataset would entail higher computational times, or costs in the case of OpenAI models, we chose to do it for a subset of the original dataset. For each of the 51 questions included in the dataset we sampled just one answer from each of the three grading levels to be rated by the models. This novel sub-sampled dataset consisted of 153 question-answer pairs, for which we ran the experiments 10 times for each combination of \glspl{llm} and prompts.

In this case, our main interest was not only focused on analyzing the accuracy on each experiment (i.e. how many outputs matched with the ground truth rating for the corresponding answer) but also the repeatability of the results obtained. For example, we could have a very accurate model, but it could be inconsistent if the errors it makes are not always with the same answers. On the other hand, a model could be not so accurate but highly consistent, if the errors it makes always correspond to the same answers. However, the main insight from the experiments carried out is that there is a high correlation between the performance of the models and their consistency.

\begin{figure}
    \centering
    \includegraphics[width=\linewidth]{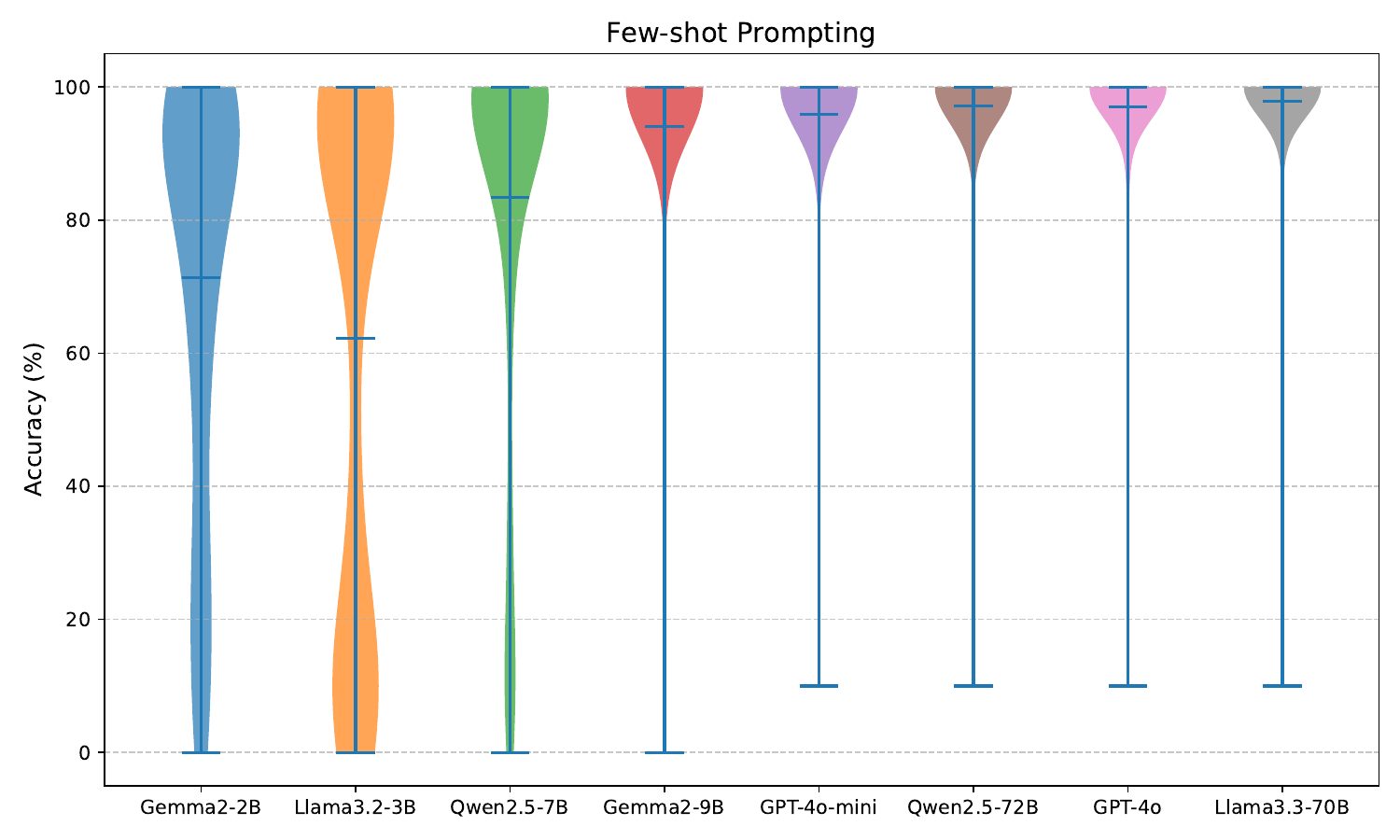}
    \caption{Accuracy distribution among the different question-answer pairs, for the best results of each model based on few-shot prompting.}    
    \label{fig:consistency}
\end{figure}

As an example of the experiment results, Figure~\ref{fig:consistency} illustrates the accuracy distribution for all the models using the few-shot based prompting strategy. The accumulation at 100\% indicates the cases that are always correctly classified, while the opposite occurs at 0\%, where the cases that are always incorrectly classified in the 10 executions fall. As we can see, all the models with an average accuracy over 90\%, also present a high consistency, with distributions that are skewed towards 100\%. It is interesting to note that the best models have no cases at 0\%, which indicates that all of the answers are rated correctly at least once.

Based on the results obtained, we can say that the best combinations of models and prompts have a very good consistency. This is extremely important and paves the way to the possibility of incorporating this type of solutions for automated task grading. Furthermore, the possibility of further regulating this variability through parameters such as model temperature still leaves room for possible improvement to obtain even more consistent results.

\subsection{Deep Inspection of the Results}

Finally, the last analysis carried out corresponds to a deeper inspection of the grades assigned by the models. In particular, we would like to know whether the errors committed are between close ratings or not. In terms of the education application, it is not the same to assign an acceptable to an answer that should be graded as excellent, that grading the same answer as incorrect. For this purpose we generated the confusion matrices for all cases, presenting here the most relevant ones.

\begin{figure}
    \centering
    \begin{subfigure}[c]{0.5\textwidth}
        \caption{GPT-4o}
        \includegraphics[width=\linewidth]{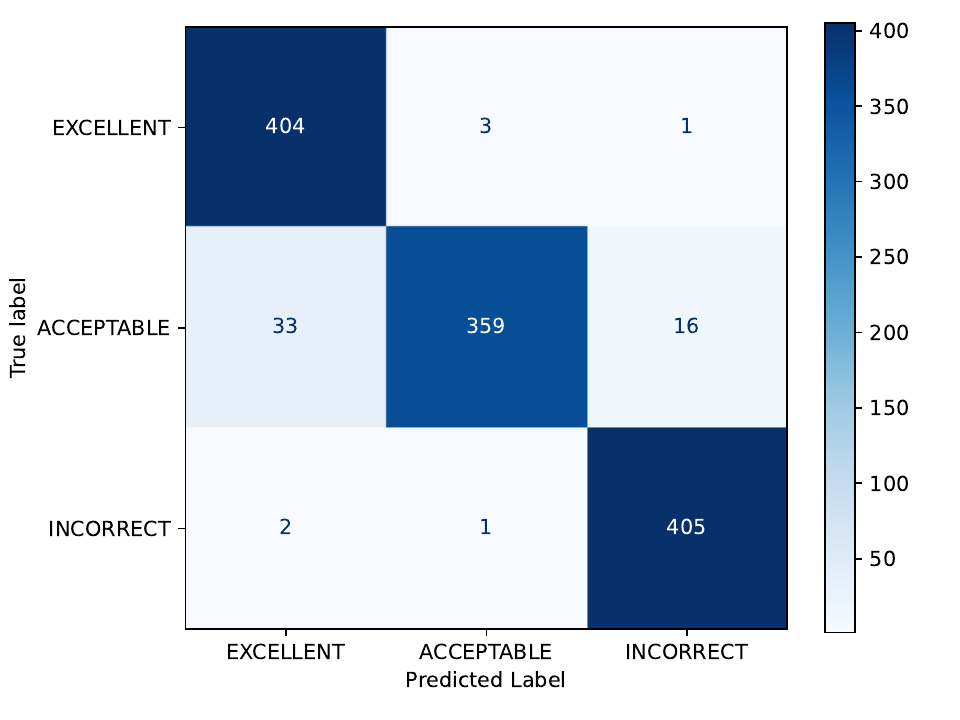}
        \label{fig:gpt4o_matrix}
    \end{subfigure}\hfill
    \begin{subfigure}[c]{0.5\textwidth}
        \caption{Llama3.3-70B}
        \includegraphics[width=\linewidth]{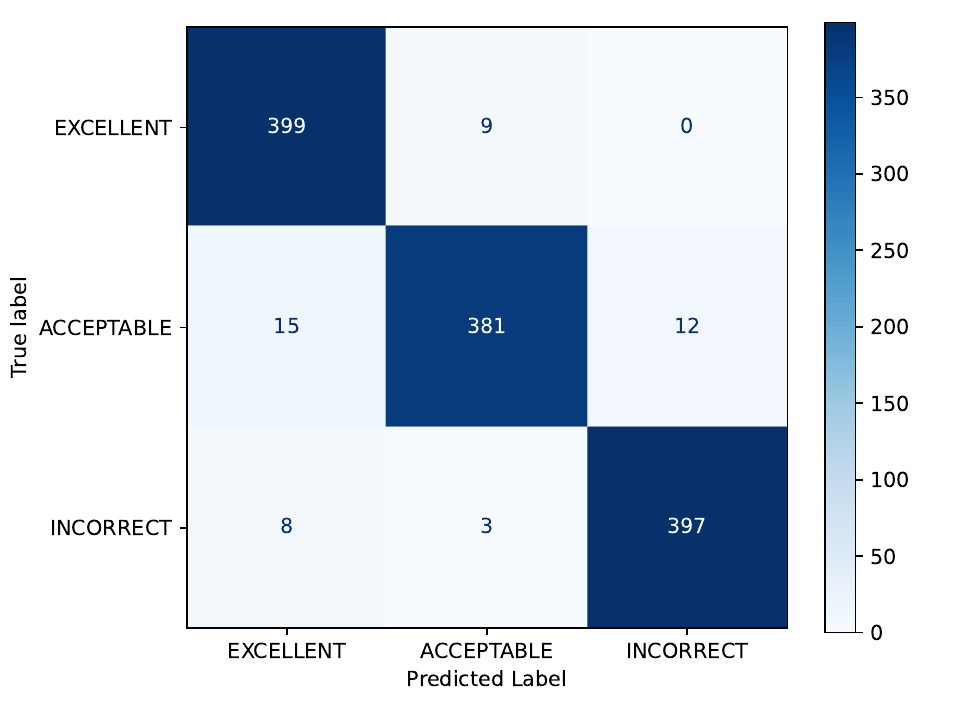}
        \label{fig:llama3.3_matrix}
    \end{subfigure}
    \caption{Confusion matrices for the best performance results: GPT-4o and Llama3.3-70B with the few-shot prompting strategy.}
    \label{fig:best_matrix}
\end{figure}

Figure~\ref{fig:gpt4o_matrix} shows the confusion matrix for the best result for GPT-4o using the few-shot prompting strategy.
It can be seen that most of the errors correspond to acceptable answers that are rated as excellent. There are also some errors that correspond to acceptable answers that the model grades as incorrect. If we simplify the task to a binary classification between correct and incorrect answers, considering both excellent and acceptable as correct answers, the model performance boosts up to 98.5\%.

Concerning the best result for the Llama3.3-70B open model, in
Figure~\ref{fig:llama3.3_matrix} we can see the confusion matrix using the few-shot prompting strategy. In this case we observe some misconfusions between the acceptable and excellent ratings and vice versa. It is observed again, that by simplifying the task to a right or wrong binary grading, the performance also increases significantly, reaching an accuracy of 98.8\%.

\begin{figure}
    \centering
    \begin{subfigure}[c]{0.5\textwidth}
        \caption{Llama3.2-3B}
        \includegraphics[width=\linewidth]{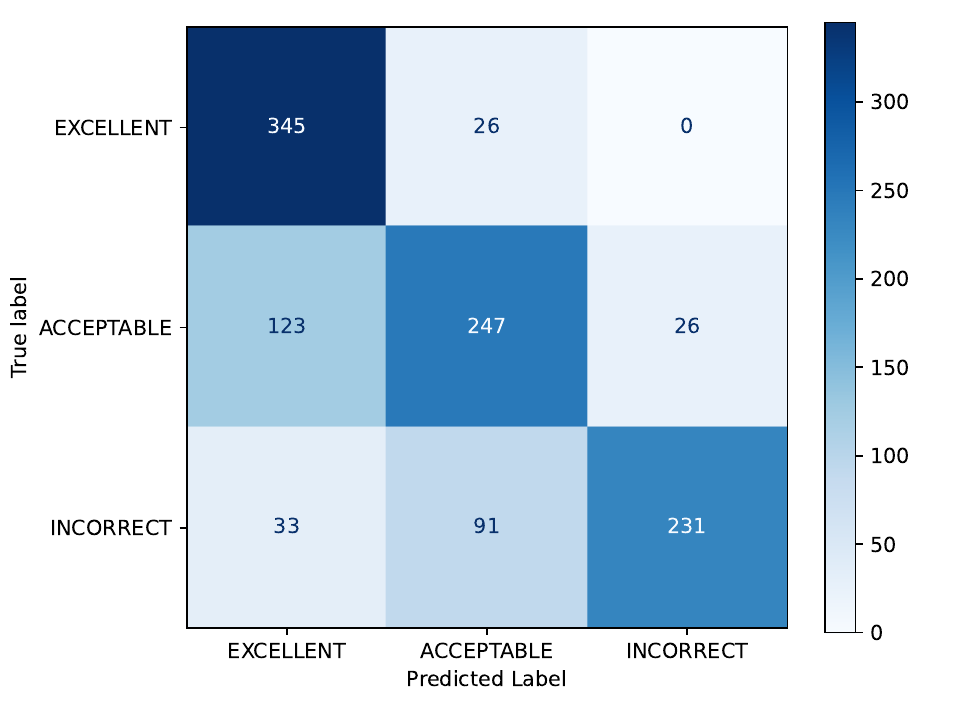}
        \label{fig:llama3.2_matrix}
    \end{subfigure}\hfill
    \begin{subfigure}[c]{0.5\textwidth}
        \caption{Gemma2-9B}
        \includegraphics[width=\linewidth]{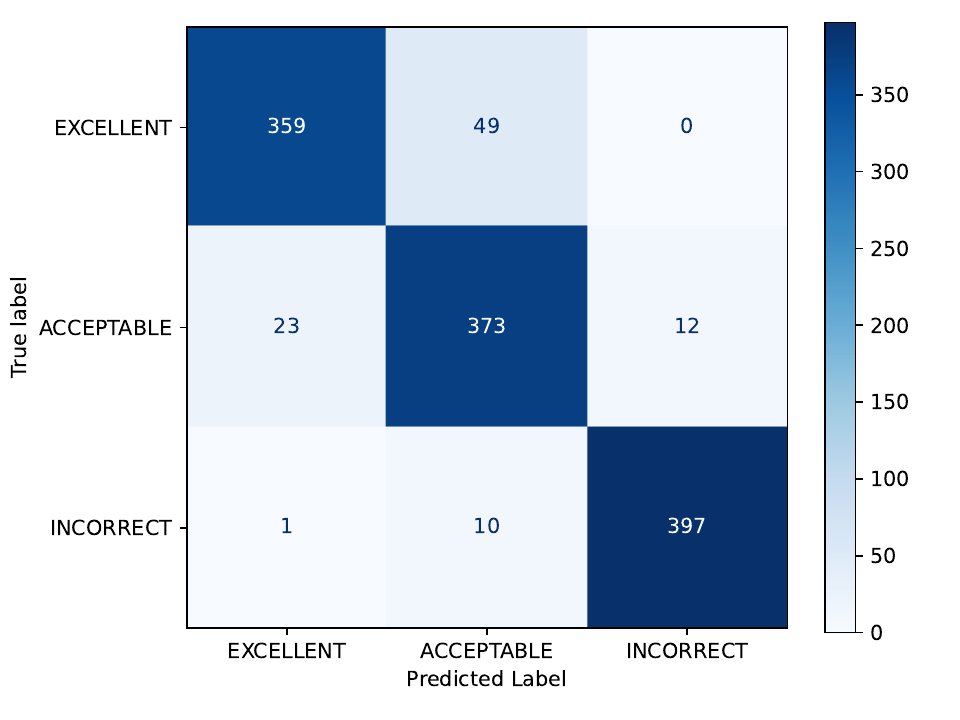}
        \label{fig:gemma_matrix}
    \end{subfigure}
    \caption{Confusion matrices for the best results with small and medium-sized models: Llama3.2-70B and Gemma2-9B with the few-shot prompting strategy.}
    \label{fig:small_matrix}
\end{figure}

Finally, the best results for small and medium-sized models are analyzed. Figures~\ref{fig:llama3.2_matrix} and~\ref{fig:gemma_matrix} presents the confusion matrices for the Llama3.2-3B and Gemma2-9B respectively, both of them using the few-shot prompting strategy. We can see that the smaller Llama3.2-3B model tends to over-rate, with many acceptable errors being assigned as excellent, as well as incorrect ones being rated as acceptable. On the other hand, Gemma2-B errors seem to be more randomly distributed without a clear pattern. However, when considering the simplified binary grading problem, this model achieves an excellent accuracy of 98.2\%, which is a great result considering the size of this model. The results for Llama3.2-3B are also not bad at all concerning the size, reaching an 87.7\% for the right vs wrong grading task.

Regarding the previous prompt sensitivity analysis presented in subsection~\ref{sec:prompt_sens}, we also compared the results with this subtle change in the prompt, for the simplified binary grading task. We observed again a notorious impact in the models performance. To illustrate the impact with some numbers, in the three best cases based on few-shot prompting, the performance drops are: from 98.8\% to 95.9\% for Llama3.3-70B, 98.5\% to 95.8\% for GPT-4o, and 98.2\% to 94.9\% Gemma2-9B. This again reinforces the idea of the great relevance that the particular words used in the prompts has, since even very subtle changes can generate very significant performance impact, as we saw in the experiments carried out.

%% file: ethics_conc_fw.tex
\section{Conclusions and Future Work}\label{sec:concl_and_fw}

This work represents a pioneering effort in the domain of automated grading systems tailored specifically for the Spanish language. By focusing on a dataset of open-ended questions and answers in Spanish, as well as evaluating state-of-the-art prompting techniques, we have provided valuable insights into the potential and current limitations of large language models (LLMs) in educational applications.

The comprehensive comparison of various LLMs and prompting strategies revealed that the current generation of language models demonstrates sufficient capability for implementing automated grading solutions. Notably, the study achieved high levels of accuracy, consistently exceeding 95\% for a three-level grading task (excellent, acceptable, and incorrect). Moreover, accuracy rose to nearly 99\% for the simplified two-level task, which involved classifying responses as either correct or incorrect. These findings underscore the robustness of LLMs in handling nuanced grading tasks when equipped with well-designed prompts.

One key observation was the sensitivity of LLMs to specific phrasing within prompts. Despite this, the adoption of advanced prompting strategies such as few-shot prompting, proved effective in mitigating this sensitivity and enhancing performance. These results suggest that prompt design remains a critical area for achieving optimal outcomes in automated grading systems.

Looking ahead, this work opens several avenues for future exploration. First, the consistency of LLM outputs could be further analyzed by varying the temperature parameter, which influences the randomness of model responses. Such an analysis would help determine the impact of temperature on grading consistency and overall accuracy.

Additionally, extending the sensitivity analysis to include modifications in student responses without altering their semantic meaning could provide deeper insights. Examining whether LLMs can maintain consistent grading levels despite minor lexical changes would be crucial for assessing their robustness in diverse classroom scenarios.

Another critical area for future research involves comparing the performance of LLMs with that of human teachers. Designing experiments to measure inter-rater reliability between automated systems and educators would not only help establish a benchmark for accuracy but also illuminate the \emph{glass ceiling} of performance achievable through automated methods. Collaborative work with teachers could also explore the use of variable prompts, wherein educators define task-specific prompts, enabling an evaluation of their impact on grading outcomes.

Finally, transitioning these solutions into production environments necessitates a thorough ethical examination. It is imperative to consider the broader implications of incorporating automated grading systems in education, including their impact on teaching practices, learning experiences, and equity. Engaging stakeholders in these discussions will be essential to ensure responsible and effective integration of this technology into educational contexts. By addressing these challenges and opportunities, future work can further advance the field of automated grading, paving the way for innovative and equitable applications in education.

\section{Ethical Considerations}\label{sec:ethics}
This article focuses on the effectiveness of using \glspl{llm} for automating the grading of short answers to open-ended questions in Spanish. However, the task itself must be evaluated in a holistic manner. Indeed, any introduction of \gls{genai} into an educational context must be carefully studied, considering not only technical aspects but also ethical and pedagogical ones. This consideration becomes even more critical in related use cases, such as providing feedback or grading other types of content generated by students.

On the one hand, the automation process requires further research to evaluate the biases of the models regarding content, wording, and the register of the answers. It also demands a robust assessment of the fairness of the corrections across different students' responses to the same question, as  previously discussed in Section~\ref{sec:concl_and_fw} as a posible line of future research work.

On the other hand, pedagogical aspects, such as the effectiveness of automatically generated feedback compared to feedback from a real educator, remain unclear. Further research in the field of behavioral economics is necessary. Additionally, the role of educators in the grading process, the accountability of the scores and feedback provided, and the consideration of contextual information, such as the students' level, must be thoroughly addressed.

All in all, these and other factors must be carefully considered, and human-in-the-loop solutions might be implemented to ensure these aspects are appropriately addressed.

%% file: prompts.tex
\section{Detailed Prompts Used in the Experiments}
\label{sec:app_prompts}

In this appendix we include all the prompts used in the experiments, which can also be found in the corresponding article repository available on GitHub: \textcolor{gray}{Link omitted for double-blind review}. 
It is worth noting that all the promtps include the variables (indicated between brackets following the promptfoo syntax). Two differente variables are used throughout the prompts, one for the dataset questions (\emph{pregunta}) and another one for the student's responses (\emph{respuesta}).

\noindent

\begin{center}
\fcolorbox{black}{lightgray}{\parbox{0.95\linewidth}{
{\bf Short Concise Prompt}\\
``Debes evaluar una respuesta a la pregunta "\{\{pregunta\}\}". La escala de evaluación tiene tres niveles: EXCELENTE, ACEPTABLE, INCORRECTA. La respuesta a la pregunta "\{\{pregunta\}\}" que tienes que evaluar es: "\{\{respuesta\}\}" Te pido que califiques la respuesta. La salida debe ser "Calificación: X", siendo X una de las opciones de calificación indicadas anteriormente.''}}
\end{center}

\begin{center}
\fcolorbox{black}{lightgray}{\parbox{0.95\linewidth}{
{\bf Long Detailed Prompt}\\
``Eres docente de educación primaria y debes evaluar una respuesta a la pregunta "\{\{pregunta\}\}" realizada a escolares entre 8 y 10 años de edad. Deberás evaluar la respuesta, teniendo en cuenta su completitud, nivel de detalle y justificación, y si está adecuada a la consigna planteada. La escala de evaluación tiene tres niveles: EXCELENTE, ACEPTABLE, INCORRECTA. EXCELENTE corresponde a una respuesta muy completa, bien justificada y con detalles relevantes. ACEPTABLE corresponde a respuestas que estén correctas, es decir que se ajustan a la pregunta, pero no dan mucha justificación ni detalle. INCORRECTA corresponde a respuestas incorrectas, que no cumplen con la consigna y están equivocadas. La respuesta a la pregunta "\{\{pregunta\}\}" que tienes que evaluar es: "\{\{respuesta\}\}" Te pido que califiques la respuesta. La salida debe ser "Calificación: X", siendo X una de las opciones de calificación indicadas anterio
rmente.''}}\\
\end{center}

\begin{center}
\fcolorbox{black}{lightgray}{\parbox{0.95\linewidth}{
{\bf \gls{cot}-based Short Prompt}\\
``Debes evaluar una respuesta a la pregunta "\{\{pregunta\}\}". La escala de evaluación tiene tres niveles: EXCELENTE, ACEPTABLE, INCORRECTA. La respuesta a la pregunta "\{\{pregunta\}\}" que tienes que evaluar es: "\{\{respuesta\}\}" Te pido que des una devolución y califiques la respuesta. La salida debe ser "Devolución: texto\_devolucion. Calificación: X", siendo texto\_devolucion los comentarios sobre la respuesta y X una de las opciones de calificación indicadas anteriormente.''}}\\
\end{center}

\begin{center}
\fcolorbox{black}{lightgray}{\parbox{0.95\linewidth}{
{\bf \gls{cot}-based Long Prompt}\\
``Eres docente de educación primaria y debes evaluar una respuesta a la pregunta "\{\{pregunta\}\}" realizada a escolares entre 8 y 10 años de edad. Deberás evaluar la respuesta, teniendo en cuenta su completitud, nivel de detalle y justificación, y si está adecuada a la consigna planteada. La escala de evaluación tiene tres niveles: EXCELENTE, ACEPTABLE, INCORRECTA. EXCELENTE corresponde a una respuesta muy completa, bien justificada y con detalles relevantes. ACEPTABLE corresponde a respuestas que estén correctas, es decir que se ajustan a la pregunta, pero no dan mucha justificación ni detalle. INCORRECTA corresponde a respuestas incorrectas, que no cumplen con la consigna y están equivocadas. La respuesta a la pregunta "\{\{pregunta\}\}" que tienes que evaluar es: "\{\{respuesta\}\}" Te pido que des una devolución y califiques la respuesta. La salida debe ser "Devolución: texto\_devolucion. Calificación: X", siendo texto\_devolucion los comentarios sobre la respuesta y X una de las opciones de calificación indicadas anteriormente.''}}\\
\end{center}

\begin{center}
\fcolorbox{black}{lightgray}{\parbox{0.95\linewidth}{
{\bf Few-Shot based Prompt}\\
``Debes evaluar una respuesta de escolares a una pregunta abierta. La escala de evaluación tiene tres niveles: EXCELENTE, ACEPTABLE, INCORRECTA.
Algunos ejemplos de referencia son los siguientes:\\

Pregunta: ¿En qué te pareces a tu hermano?
Respuesta: Me parezco a mi hermano en que ambos disfrutamos mucho los videojuegos; los jugamos todos los días juntos y compartimos nuestros favoritos.\\
Calificación: EXCELENTE\\

Pregunta: ¿En qué te pareces a tu hermano?
Respuesta: A ambos nos gustan los videojuegos.\\
Calificación: ACEPTABLE\\

Pregunta: ¿En qué te pareces a tu hermano?
Respuesta: Él me ayuda mucho con los estudios antes de jugar.\\
Calificación: INCORRECTA\\

En este caso debes evaluar una respuesta a la pregunta "\{\{pregunta\}\}". La respuesta que tienes que evaluar es: "\{\{respuesta\}\}"
Te pido que califiques la respuesta. La salida debe ser "Calificación: X", siendo X una de las opciones de calificación indicadas anteriormente.''}}\\
\end{center}

The modified versions used in the prompts sensibility experiments just correspond to replace the particular words used for each grading level:
\begin{itemize}
\item \texttt{SOBRESALIENTE} instead of \texttt{EXCELENTE}.
\item \texttt{BIEN} instead of \texttt{ACEPTABLE}.
\item \texttt{DEFICIENTE} instead of \texttt{INCORRECTA}.
\end{itemize}